# RAG/LLM Augmented Switching Driven Polymorphic Metaheuristic Framework

**[1]Faramarz Safi Esfahani, [1]Ghassan Beydoun,**
**[2]Morteza Saberi, [2]Brad McCusker, [4]Biswajeet Pradhan**

faramarz.safiesfahani@uts.edu.au, ghassan.beydoun@uts.edu.au, morteza.saberi@uts.edu.au,
brad.mccusker@surroundaustralia.com, biswajeet.pradhan@uts.edu.au

[1] School of Information, Systems and Modelling, Faculty of Engineering and IT, University of Technology Sydney,
Sydney, NSW, Australia.
[2] School of Computer Science and DSI, University of Technology Sydney, Australia,
Sydney, New South Wales, Australia.
[3] Surround Australia, Canberra, Australia.
[4] Centre for Advanced Modelling and Geospatial Information Systems, School of Civil and Environmental
Engineering, Faculty of Engineering and Information, Sydney, New South Wales, Australia.

**Abstract:**
Metaheuristic algorithms are widely used for solving complex optimization problems, yet their effectiveness is often constrained by fixed structures and the need for extensive tuning. The Polymorphic Metaheuristic Framework (PMF) addresses this limitation by introducing a self-adaptive metaheuristic switching mechanism driven by real-time performance feedback and dynamic algorithmic selection. PMF leverages the Polymorphic Metaheuristic Agent (PMA) and the Polymorphic Metaheuristic Selection Agent (PMSA) to dynamically select and transition between metaheuristic algorithms based on key performance indicators, ensuring continuous adaptation. This approach enhances convergence speed, adaptability, and solution quality, outperforming traditional metaheuristics in high-dimensional, dynamic, and multimodal environments. Experimental results on benchmark functions demonstrate that PMF significantly improves optimization efficiency by mitigating stagnation and balancing exploration-exploitation strategies across various problem landscapes. By integrating AI-driven decision-making and self-correcting mechanisms, PMF paves the way for scalable, intelligent, and autonomous optimization frameworks, with promising applications in engineering, logistics, and complex decision-making systems.

# Introduction

In the rapidly evolving field of optimization and artificial intelligence, traditional metaheuristic algorithms have demonstrated strong performance in solving complex problems. However, their effectiveness is often limited by fixed algorithmic structures, requiring extensive manual tuning and domain-specific knowledge to adapt to different problem landscapes(Luo et al., 2024). With the emergence of Retrieval Augmented Generative AI (RAG) and LLMs (Large Language Models) and their ability to process and synthesize vast amounts of knowledge, there is an opportunity to create a self-adaptive, intelligent optimization framework capable of dynamically leveraging metaheuristics based on real-time performance.

Metaheuristic algorithms have been widely used for solving complex optimization problems due to their ability to explore large search spaces efficiently. However, these algorithms often require manual tuning and are limited by fixed exploration-exploitation strategies, making them suboptimal in dynamic and multimodal environments. Recent research has explored adaptive metaheuristics, such as the Adaptive Metaheuristic Framework (AMF)(Ahmed, 2024), which switches strategies based on predefined rules, and reinforcement learning-based approaches, which adjust parameters dynamically. Additionally, LLM-driven approaches such as LLaMEA (Van Stein & Bäck, n.d.) and CRISPE (Zhong et al., n.d.) have introduced the idea of using AI-generated heuristics, though they remain largely focused on static optimization rather than dynamic selection and adaptation. While these approaches have contributed significantly to the field, there is still a lack of a unified framework that integrates AI decision-making with real-time algorithmic switching in a self-correcting feedback loop.

Existing metaheuristic frameworks lack adaptability when dealing with complex, dynamic, and multimodal optimization problems, such as those in recent CEC benchmarks (Luo et al., 2024), which require algorithms to track multiple optima in changing environments. While some frameworks introduce rule-based adaptation or reinforcement learning, they do not leverage the potential of LLMs for high-level decision-making. The challenge remains: how can we create a framework that autonomously selects, adapts, and transitions between different metaheuristic algorithms based on real-time performance feedback?

This research hypothesis that by dynamically selecting and switching metaheuristics based on real-time performance feedback, better fitness values, faster convergence, and improved adaptability compared to individual static metaheuristic algorithms, are achieved. The Polymorphic Metaheuristic Framework (PMF) provided by this research, addresses this gap by combining AI-driven reasoning with traditional optimization methods, allowing for dynamic metaheuristic selection, population transfer, and real-time performance tracking, making it a self-adaptive, AI-enhanced optimization framework. The Polymorphic Metaheuristic Framework (PMF) consists of two main components: 1) PMA (Polymorphic Metaheuristic Agent) – Responsible for orchestrating the utilization of metaheuristic algorithms based on the feedback it receives from the PMSA; 2) PMSA (Polymorphic Metaheuristic Selection Agent) – Evaluates the performance of the current metaheuristic and provides feedback to the PMA. The PMSA (Polymorphic Metaheuristic Selection Agent) can be implemented independently or integrated with a Large Language Model (LLM) / Retrieval-Augmented Generation (RAG) for enhanced decision-making. Based on this feedback, the PMA either continues with the current metaheuristic algorithm or switches to another one. Additionally, several strategies for population switching are introduced. The case study shows that the PMF outweighs the baseline metaheuristics individually.

From now on, the structure of the article is organized as follows: first, related works are reviewed; next, the Polymorphic Metaheuristic Framework (PMF) is introduced; finally, a case study is presented to evaluate the behavior of the proposed PMF.

# Related works

The concept of dynamically selecting and adapting metaheuristic algorithms using Large Language Models (LLMs) has been explored in recent research. Several studies have investigated similar concepts, emphasizing adaptive and flexible metaheuristic frameworks that can adjust to dynamic environments. The section concludes with a comparative analysis of related studies, summarized in Table 1.

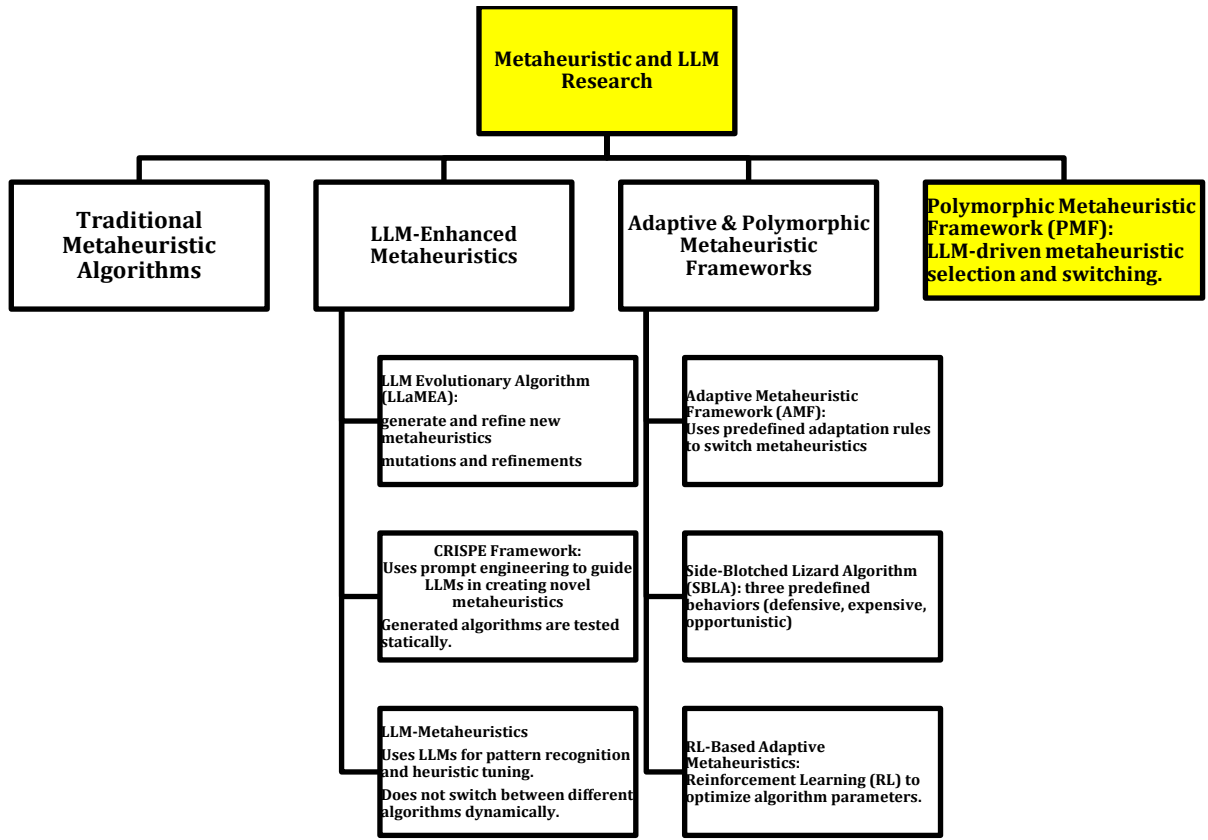

**Figure 1: The mind-map of this research.**

The research paper (Ahmed, n.d.) introduces an Adaptive Metaheuristic Framework designed to intelligently adapt to changes in problem parameters. The framework combines a dynamic representation of problems, a real-time sensing system, and adaptive techniques to navigate continuously changing optimization environments.

Authors in (Maciel C. et al., 2020) present the Side-Blotched Lizard Algorithm (SBLA), an optimization strategy designed to emulate the polymorphism observed in lizard populations. SBLA incorporates three morphs with specific behaviors—defensive, expansive, and opportunistic—to balance exploration and exploitation in the search process.

This research (Tessari & Iacca, 2022)introduces a framework that integrates reinforcement learning with metaheuristic algorithms to enable parameter adaptation. The framework is applied to algorithms like Covariance Matrix Adaptation Evolution Strategies (CMA-ES) and Differential Evolution (DE), allowing them to adjust parameters dynamically based on the problem landscape.

AutoOpt (Zhao et al., n.d.) proposes a framework for the automated design of metaheuristic algorithms with diverse structures. It introduces a general algorithm prototype and a directed acyclic graph representation to explore various algorithm structures, promoting flexibility and adaptability in optimization.

The study (Van Stein & Bäck, n.d.) introduces LLaMEA, an evolutionary algorithm that leverages LLMs to automatically generate and optimize high-quality metaheuristic optimization algorithms. The framework combines in-context learning with precise task prompts, including sample algorithms, code interface requirements, and mutation instructions.

The research (Zhong et al., n.d.) presents the CRISPE framework, which utilizes LLMs like PMSA-3.5 to design new metaheuristic algorithms. The framework employs prompt engineering to guide the LLM in generating algorithms inspired by natural phenomena, such as the proposed Zoological Search Optimization (ZSO) algorithm.

The paper (Chacón Sartori et al., n.d.) discusses a novel approach that leverages LLMs as pattern recognition tools to improve metaheuristics. It explores the integration of LLMs into metaheuristic frameworks to enhance optimization processes.

In summary, as shown in Table 1, while all these approaches aim to enhance optimization through the integration of LLMs, PMF distinguishes itself by focusing on the dynamic selection and adaptation of existing metaheuristic algorithms based on real-time feedback, offering a practical and adaptable solution. PMF is more practical for real-world applications, as it adapts existing metaheuristics rather than generating new ones.

| Feature | LLaMEA (LLM Evolutionary Algorithm) | CRISPE (LLM-Generated Metaheuristics) | LLM-Metaheuristics (LLM-Assisted Optimization) | AMF (Adaptive Metaheuristic Framework) | SBLA (Side-Blotched Lizard Algorithm) | RL-Based Adaptive Metaheuristics | AutoOpt (Automated Metaheuristic Discovery) | PMF (Polymorphic Metaheuristic Framework) Our Approach |
|---|---|---|---|---|---|---|---|---|
| Adaptability | Evolves generated algorithms iteratively | Static, generated metaheuristics | Enhances heuristics based on patterns | Adapts algorithms based on environmental changes | Uses polymorphic strategy (3 morphs) | RL model tunes parameters dynamically | Evolves new metaheuristics over time | Dynamically selects metaheuristics based on feedback |
| Algorithm Selection | Generates new algorithms rather than selecting | Uses a fixed LLM-generated heuristic | Uses LLM to improve existing heuristics | Adjusts algorithm based on performance changes | Fixed strategy (lizard morphs) | RL selects optimal strategies | Searches and refines algorithms dynamically | Uses PMSA to choose the best heuristic dynamically |
| Algorithm Adaptation | Evolves new metaheuristics iteratively | Does not adapt—static generation | Adjusts existing algorithms | Adjusts parameters dynamically | Predefined adaptation rules | Learns adaptation policy through RL | Can adjust parameters or design new algorithms | Switches to different algorithms based on feedback |
| Algorithm Generation | Generates brand-new metaheuristics | Generates a new algorithm from scratch | No generation, just refines heuristics | Does not generate new algorithms | Uses pre-set morph strategies | Modifies existing metaheuristics | Can construct novel metaheuristic algorithms | Fetches or switches to existing algorithms |
| Feedback Mechanism | Uses evolutionary learning for refinement | No feedback mechanism—static design | Uses LLM to refine heuristics | Monitors environmental shifts for adaptation | No explicit feedback—fixed morph behaviors | RL continuously learns from results | Uses optimization performance to evolve algorithms | Uses performance metrics to modify selection |
| Computational Complexity | High—evolutionary algorithm generation is costly | High—LLM generates and tests heuristics | Moderate—depends on LLM efficiency | Requires problem-specific adaptation rules | Simple—uses fixed morph strategies | High—requires RL training | High—requires extensive automated search | Moderate—requires API calls and script management |
| Diversity of Algorithms | Can generate unlimited metaheuristics | Generates new but limited heuristic classes | Improves existing heuristics but does not create new ones | Typically focuses on one algorithm per environment | Includes three morphs but is fixed per scenario | Learns only one algorithm at a time | Evolves entirely new structures | Supports multiple known metaheuristics |
| Optimization Domains | General-purpose, depends on training | Focused on nature-inspired optimization | Enhances existing optimization techniques | Adaptive across changing environments | Evolutionary population-based optimization | Works in structured optimization problems | Generalized across multiple problems | General-purpose, problem-independent |
| Implementation Complexity | High—requires evolving new metaheuristics | High—depends on prompt engineering | Moderate—uses LLM refinements | Requires custom adaptation rules | Simple, predefined heuristics | Requires extensive RL training | High—requires automated algorithm discovery | Requires API calls and script handling |

Table 1: Comparison of PMA with Related Approaches

## Polymorphism Metaheuristic Framework

The Polymorphic Metaheuristic Framework (PMF) illustrated in Figure 1 is an adaptive, AI-enhanced metaheuristic optimization system that leverages Polymorphic Metaheuristic Selection Agent (PMSA) to select a right metaheuristic algorithm based on the runtime circumstances. The PMSA can be realized by applying both Retrieval Augmented Generative (RAG) and Large Language Model (LLM) - RAG/LLM - driven decision-making to dynamically select and switch between different optimization algorithms based on real-time performance feedback. Unlike traditional metaheuristic approaches that rely on fixed strategies, PMF acts as an intelligent orchestration layer, continuously evaluating algorithm effectiveness using performance metrics such as fitness scores, convergence rates, and exploration-exploitation balance. PMF includes two major counterparts called Polymorphic Metaheuristic Agent (PMA), and PMSA. Through an interactive feedback loop with PMSA, PMF determines whether to continue with the current algorithm or switch to a more suitable alternative, ensuring robust adaptability across diverse problem landscapes. This framework allows to operate on dynamic multimodal optimization problems, where multiple optima and changing environments require continuous adaptation. Additionally, PMF supports seamless population transfer mechanisms, maintaining search efficiency when switching between metaheuristics. By combining metaheuristic flexibility with LLM intelligence, PMA represents a hybrid framework capable of bridging the gap between static heuristics and AI-driven optimization, offering a scalable and self-corrective approach to solving complex optimization problems.

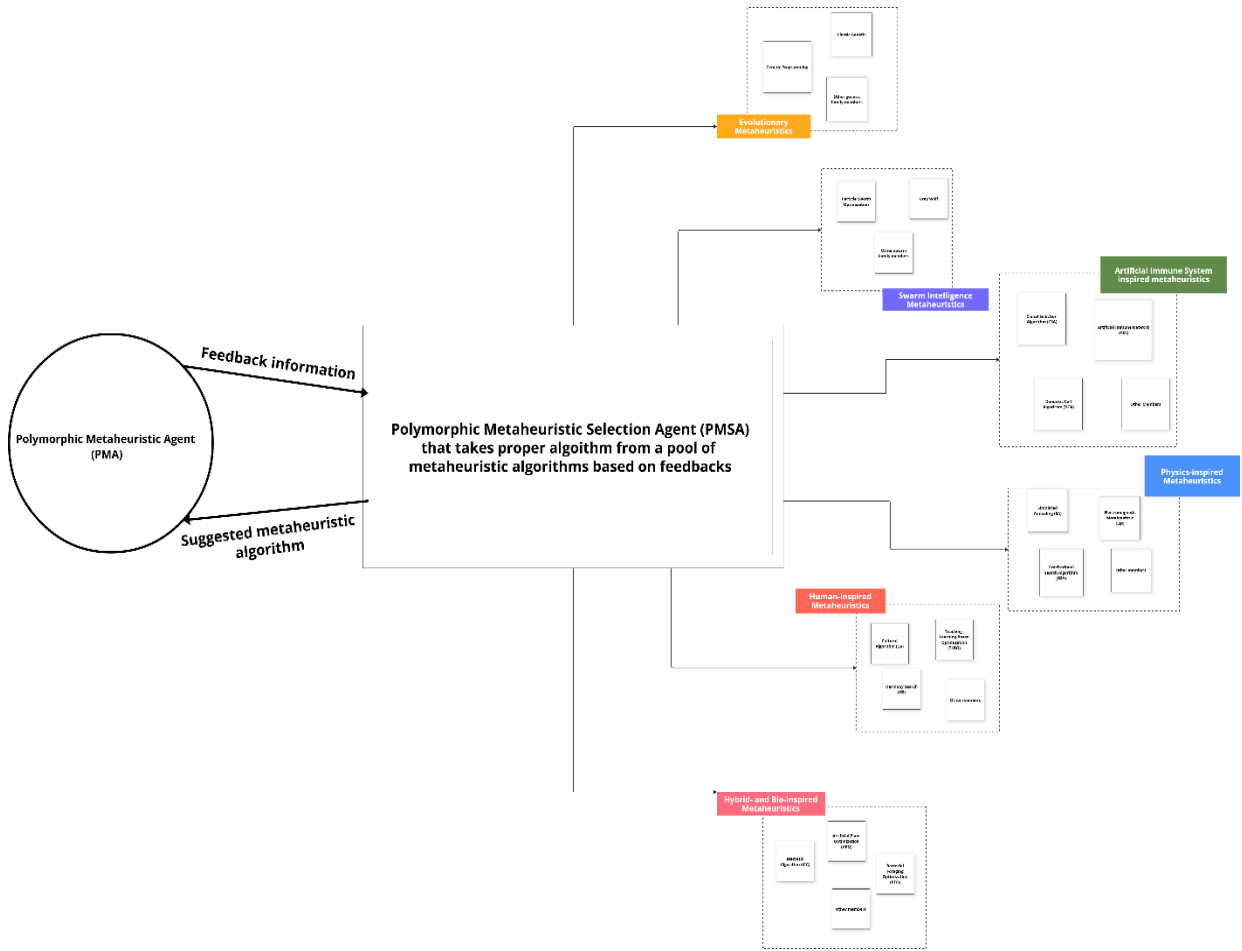

**Figure 2: Polymorphic Metaheuristic Framework (PMF)**

## *Polymorphic Metaheuristic PMA and PMSA*

This section breaks down the decision-making process used by the PMSA and possible reasons for its behavior. The PMSA follows a decision process based on the feedback parameters provided by PMA. Below is the expected reasoning that PMSA uses when responding to the PMA agent.

## *Feedback structure*

The feedback loop in PMF is designed to ensure continuous adaptation and optimization by dynamically selecting the most suitable metaheuristic algorithm based on real-time performance. At each iteration, PMA evaluates the selected algorithm's effectiveness using key performance indicators such as fitness value, convergence rate, solution improvement rate, stagnation count, and exploration-exploitation balance. This feedback is then logged and analyzed, and a structured query is sent to PMSA, asking whether to continue with the current algorithm or switch to another. If a switch is recommended, PMA adapts the population by transferring the best solutions from the previous algorithm to the newly selected one, ensuring continuity in the search process. This loop repeats for every iteration, allowing PMA to dynamically adjust its strategy in response to changing optimization landscapes, making it particularly well-suited for complex, dynamic, and even multimodal problems. The continuous feedback and adaptation mechanism give PMA a self-correcting ability, bridging the gap between traditional static heuristics and truly adaptive, AI-enhanced metaheuristic frameworks. To recap the feedback loop improves PMA's Decision Making by 1) Tracking whether the current algorithm is stagnating, allowing PMSA to detect when a switch is needed. 2) Balancing exploration vs. exploitation, so PMSA can adjust for more diverse searching or focused refining. 3) Logging historical performance trends, helping PMSA detect which algorithm has worked best in past iterations. 4) Monitoring computational cost, so PMSA can avoid expensive algorithms if needed.

### *Population Transfer Between Iterations*

In metaheuristic optimization, the way a population (set of candidate solutions) transitions from one iteration to the next is crucial for ensuring efficient exploration and exploitation of the search space. PMA ensures a seamless population transition between algorithms, preventing loss of good solutions that should be managed across iterations.

Algorithm Selection: If PMSA recommends the same algorithm, it runs normally. If PMSA switches to a new algorithm, population adaptation occurs. Elite Preservation: Best solutions from the previous iteration are preserved (e.g., top 10% of solutions). Population Mapping (if needed): If the new algorithm uses a different representation, the solutions are transformed accordingly. Diversity Check & Restart: If population diversity is too low after switching, a restart mechanism ensures better exploration. Re-Evaluation & Initialization: The new algorithm re-evaluates the imported solutions, adjusting parameters dynamically.

### *Population Handover Strategy*

Since PMA dynamically selects and switches between algorithms, the population transfer mechanism may vary depending on whether: The same algorithm continues into the next iteration, or a new algorithm is selected, requiring population adaptation. If PMF decides to switch to another algorithm, a population handover strategy is required to avoid losing valuable information. The transition involves:

Direct Population Injection: The best-performing individuals from the previous algorithm are directly injected into the new algorithm. This helps the new algorithm start with promising solutions rather than random initialization. Population Re-Evaluation: If the new algorithm has different encoding requirements, solutions are re-evaluated and transformed accordingly. Hybridization Mechanism: Some algorithms combine multiple heuristics (e.g., CMA-ES + DE). PMA can merge best individuals from both algorithms to create a hybrid population.

## Case Study Results and Discussion

The proposed PMF was evaluated against seven baseline algorithms: Genetic Algorithm (GA) (Holland, 1992), Particle Swarm Optimization (PSO)(Kennedy & Eberhart, n.d.) , Differential Evolution (DE) (Storn & Price, 1997), Ant Colony Optimization (ACO) (Tan & Shi, 2024), Simulated Annealing (SA) (Kirkpatrick et al., 1983), Tabu Search (TS) (Glover, 1989), and Covariance Matrix Adaptation Evolution Strategy (CMA-ES) (Hansen & Ostermeier, 2001). The experiments were conducted using the F12022 benchmark function from the CEC2022 competition (Luo et al., 2022), with a 10-dimensional search space. Performance evaluation was conducted based on three key metrics: Fitness Value: The final optimized objective function value. Algorithm Switching Frequency: The number of times PMA switched between different algorithms. Convergence Rate: The speed at which an algorithm approaches an optimal solution.

According to Figure 3, PMA achieved a final fitness value of 10,954.30, significantly outperforming all baseline algorithms. By contrast, the best-performing baseline (PSO) converged to a suboptimal solution with a final fitness of 14,254.40, while GA stagnated at 14,007.14. PMA demonstrated a superior convergence rate, rapidly reducing the fitness from 25,044.30 in the first iteration to 10,954.30 by iteration 9, whereas DE and PSO exhibited slow and inconsistent convergence. The adaptive switching allowed PMA to transition from DE to GA, CMA-ES, SA, and ultimately ACO, ensuring continuous improvement and avoiding premature stagnation. Unlike fixed heuristic strategies, PMA effectively adapts to different optimization stages, leveraging exploratory algorithms (e.g., DE, PSO) in early iterations and switching to exploitative methods (e.g., ACO, SA) in later stages. This flexibility enhances convergence rates and prevents local minima trapping according to Figure 4.

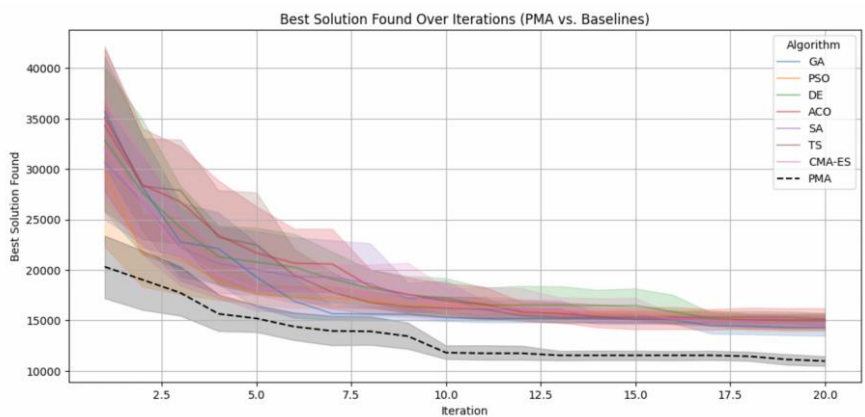

**Figure 3: Best Solution Found Over Iterations**

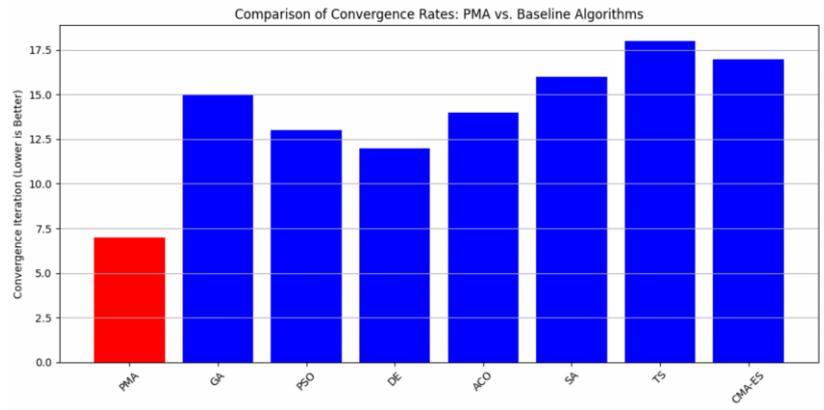

**Figure 4: Comparison of Convergence Rate vs. Baseline Algorithms**

An important advantage of PMA is its ability to dynamically switch algorithms based on performance feedback. Figure 5 illustrates the number of algorithm transitions per iteration. In Run 5, PMA switched 15 times, adapting to problem complexity and maintaining consistent progress. The most frequently selected algorithms were ACO and SA, which dominated the final iterations, contributing to fine-tuned exploitation.

The baseline algorithms, in contrast, remained fixed throughout the optimization process, GA and DE exhibited stagnation beyond iteration 10, leading to plateauing fitness improvements.

One of the major limitations of baseline algorithms is their susceptibility to stagnation. PMA effectively detects stagnation patterns and switches algorithms when solution improvement plateaus. This adaptability is crucial for maintaining continuous fitness enhancement.

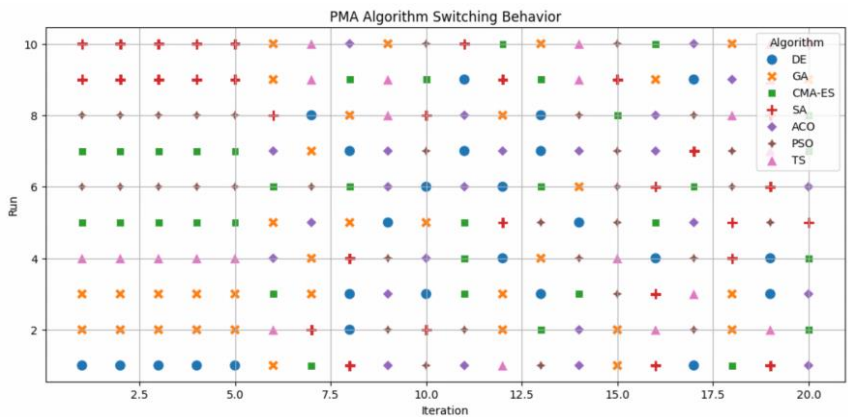

**Figure 5: PMA Algorithm Switching Behavior**

## Conclusion and Future Research

The Polymorphic Metaheuristic Framework (PMF) introduces an adaptive, AI-enhanced approach to metaheuristic optimization, dynamically selecting and switching algorithms based on real-time performance feedback. By integrating the Polymorphic Metaheuristic Agent (PMA) and the Polymorphic Metaheuristic Selection Agent (PMSA), PMF effectively balances exploration and exploitation, mitigating stagnation while improving convergence speed and solution quality. Experimental results demonstrate that PMF outperforms traditional static metaheuristics by continuously adapting to changing optimization landscapes. Its flexibility and self-correcting feedback loop make it particularly suited for dynamic, multimodal problems, bridging the gap between classical heuristic strategies and AI-driven decision-making. The framework's success highlights its potential for broader applications in complex real-world optimization tasks, setting the stage for further advancements in adaptive metaheuristics.

While PMF demonstrates strong adaptability and superior optimization performance, several areas have been focused for future exploration: 1) Machine Learning-Driven Metaheuristic Selection – Incorporating reinforcement learning (RL) or meta-learning could refine PMSA's decision-making, allowing it to learn optimal switching strategies based on historical performance and problem-specific characteristics; 2) Multi-Objective Optimization and Real-World Applications – Extending PMF to handle multi-objective optimization (MOO) problems would improve its applicability to real-world scenarios, such as finance, logistics, and bioinformatics, where trade-offs between conflicting objectives must be managed dynamically; 3) Hybridization with Ensemble Learning – Combining PMF with ensemble learning techniques could enhance robustness, enabling the framework to generate metaheuristic ensembles that adaptively combine multiple algorithms for better optimization efficiency; 4) Scalability for Large-Scale Optimization – Investigating parallel computing and distributed optimization strategies could allow PMF to scale effectively for high-dimensional and large-scale optimization challenges; 5) Explainability and Interpretability in Metaheuristic Switching – Enhancing PMSA with explainable AI (XAI) techniques could improve transparency, providing insights into why specific metaheuristic transitions occur, thereby increasing trust in AI-driven optimization; 6) Integration with Quantum and Bio-Inspired Computation – Exploring how PMF can leverage quantum computing or bio-inspired metaheuristics could unlock new optimization capabilities, particularly in solving complex combinatorial and high-dimensional problems.

# References

Ahmed, B. S. (n.d.). *An Adaptive Metaheuristic Framework for Changing Environments*.

Ahmed, B. S. (2024). *An Adaptive Metaheuristic Framework for Changing Environments*. http://arxiv.org/abs/2404.12185

Chacón Sartori, C., Blum, C., Bistaffa, F., & Rodríguez Corominas, G. (n.d.). *Metaheuristics and Large Language Models Join Forces: Toward an Integrated Optimization Approach*. https://doi.org/10.1109/ACCESS.2024.3524176

Glover, F. (1989). Tabu Search—Part I. *Https://Doi.Org/10.1287/Ijoc.1.3.190*, *1*(3), 190–206. https://doi.org/10.1287/IJOC.1.3.190

Hansen, N., & Ostermeier, A. (2001). Completely Derandomized Self-Adaptation in Evolution Strategies. *Evolutionary Computation*, *9*(2), 159–195. https://doi.org/10.1162/106365601750190398

Holland, J. H. (1992). *[58] [Bradford Books] John H. Holland - Adaptation in Natural and Artificial Systems_ An Introductory Analysis with Applications to Biology, Control, and Artificial Intelligence (A Bradford Book) (1992, The MIT Press)*. https://books.google.com.br/books?id=JE5RAAAAMAAJ

Kennedy, J., & Eberhart, R. (n.d.). Particle swarm optimization. *Proceedings of ICNN'95 - International Conference on Neural Networks*, *4*, 1942–1948. https://doi.org/10.1109/ICNN.1995.488968

Kirkpatrick, S., Gelatt, C. D., & Vecchi, M. P. (1983). Optimization by Simulated Annealing. *Science*, *220*(4598), 671–680. https://doi.org/10.1126/SCIENCE.220.4598.671

Luo, W., Lin, X., Li, C., Yang, S., & Shi, Y. (2022). *Benchmark Functions for CEC 2022 Competition on Seeking Multiple Optima in Dynamic Environments*. http://arxiv.org/abs/2201.00523

Luo, W., Xu, P., Yang, S., & Shi, Y. (2024). *Benchmark for CEC 2024 Competition on Multiparty Multiobjective Optimization*.

Maciel C., O., Cuevas, E., Navarro, M. A., Zaldívar, D., & Hinojosa, S. (2020). Side-Blotched Lizard Algorithm: A polymorphic population approach. *Applied Soft Computing*, *88*, 106039. https://doi.org/10.1016/J.ASOC.2019.106039

Storn, R., & Price, K. (1997). Differential Evolution - A Simple and Efficient Heuristic for Global Optimization over Continuous Spaces. *Journal of Global Optimization*, *11*(4), 341–359. https://doi.org/10.1023/A:1008202821328/METRICS

Tan, Y., & Shi, Y. (Eds.). (2024). *Advances in Swarm Intelligence. 14789*. https://doi.org/10.1007/978-981-97-7184-4

Tessari, M., & Iacca, G. (2022). *Reinforcement learning based adaptive metaheuristics*. https://doi.org/10.1145/3520304.3533983

Van Stein, N., & Bäck, T. (n.d.). *LLaMEA: A Large Language Model Evolutionary Algorithm for Automatically Generating Metaheuristics*.

Zhao, Q., Yan, B., Chen, X., Hu, T., Cheng, S., & Shi, Y. (n.d.). *AutoOpt: A General Framework for Automatically Designing Metaheuristic Optimization Algorithms with Diverse Structures*. Retrieved February 9, 2025, from https://github.com/qz89/AutoOpt.

Zhong, R., Xu, Y., Zhang, • Chao, & Yu, J. (n.d.). Leveraging large language model to generate a novel metaheuristic algorithm with CRISPE framework. *Cluster Computing*, *27*. https://doi.org/10.1007/s10586-024-04654-6